\documentclass[conference]{IEEEtran}
\usepackage{multirow}
\usepackage[dvips]{graphicx}
\usepackage{amsmath}
\usepackage[psamsfonts]{amssymb}
\usepackage{amsxtra}
\usepackage{graphicx,color}
\usepackage{threeparttable}
\usepackage{url}
\usepackage{CJK}
\usepackage[utf8]{inputenc}

\usepackage[hang]{footmisc}

\begin{document}
\title{OC16-CE80: A Chinese-English Mixlingual Database and A Speech Recognition Baseline}

\author{%
\IEEEauthorblockN{%
Dong Wang\IEEEauthorrefmark{1},
Zhiyuan Tang\IEEEauthorrefmark{2},
Difei Tang\IEEEauthorrefmark{3} and
Qing Chen\IEEEauthorrefmark{3}
}

\IEEEauthorblockA{%
\IEEEauthorrefmark{1}
Center for Speech and Language Technologies, Division of Technical Innovation and Development, \\
Tsinghua National Laboratory for Information Science and Technology\\
Center for Speech and Language Technologies, Research Institute of Information Technology, Tsinghua University \\
Corresponding Author: wangdong99@mails.tsinghua.edu.cn}

\IEEEauthorblockA{%
\IEEEauthorrefmark{2}
Chengdu Institute of Computer Applications, Chinese Academy of Sciences \\
E-mail: tangzy@cslt.riit.tsinghua.edu.cn}

\IEEEauthorblockA{%
\IEEEauthorrefmark{3}
SpeechOcean, Beijing, China\\
E-mail: \{tangdifei, chenqing\}@speechocean.com}

}

\maketitle

\begin{abstract}
  We present the OC16-CE80 Chinese-English mixlingual speech database which was released
  as a main resource for training, development and test
  for the Chinese-English mixlingual speech recognition (MixASR-CHEN) challenge
  on O-COCOSDA 2016. This database consists of 80 hours of speech signals recorded from more than 1,400 speakers,
  where the utterances are in Chinese but each involves one or several English words.
  Based on the database and another two free data resources (THCHS30 and the CMU dictionary),
  a speech recognition (ASR) baseline was constructed with
  the deep neural network-hidden Markov model (DNN-HMM) hybrid system.
  We then report the baseline results following the MixASR-CHEN evaluation rules and
  demonstrate that OC16-CE80 is a reasonable data resource for mixlingual research.
\end{abstract}


\IEEEpeerreviewmaketitle

\section{Introduction}

Language is a fundamental capability of human beings. In history,
language in a particular area was fairly stable at most of the time,
which means that the evolution of language was noticeable but rather slow.
This steady evolution in history can be largely attributed to the
limited inter-area or inter-nation cultural and business interaction.

This situation has been fundamentally changed since last century.
The international business developed so rapidly that bilateral trading
quickly evolved into global trading, and national markets have become
inseparable parts of the global market. Especially in the latest twenty
years, the development of the Internet has led to much more efficient
information exchange and glued the entire earth together.
This brings about quick and remarkable
change in our daily life, particularly the trend of internationalism,
globalization and interculturalism~\cite{Baker11foundation}.

An interesting consequence of this change in human language is the
bilingual or multilingual phenomenon.
During the international interaction, different languages meet and
influence each other, leading to interesting acoustic and linguistic phenomena.
It can be simply the popularity of some foreign words, but
be more as profound as the adoption of a foreign language as
an official one~\cite{matras2009language, hickey2010handbook, bakker2013contact}.
The mutual influence among languages is universal in modern society,
e.g., the influence of Mandarin to other minor languages in China,
and the influence of English to other languages in the world.
The complex acoustic and linguistic patterns of languages in
multilingual conditions
have attracted much interest in a multitude of
research areas, including comparative phonetics, evolutionary
linguistics, language development and sociolinguistics.

Perhaps the most commonly encountered multilingual phenomenon is
mixlingual embedding, i.e., the phenomenon that one or several
words from other language
(called the secondary language or embedding language) were
embedded in the sentences of a main language (called the
primary language or matrix language).
This mixlingual problem is also called `code-switching', as it involves
the change from one language to another. This may take place at various
levels including sentence, phrase, or word~\cite{heredia2001bilingual, poplack2001code, toribio2012cambridge, auer2013code}.
For example, the following sentence involves a sentence-level code-switching,
where an English word ``Google'' is seated in the main Chinese character sequence:
``\begin{CJK}{UTF8}{gbsn}
你可以Google这篇论文
\end{CJK}''.
Speech recognition against code-switching has ignited interest
in both robust acoustic modeling~\cite{stemmer2001acoustic, lyu2006speech, vu2012first, modipa2013implications, lyudovyk2014code, yilmaz2016investigating}
and language modeling~\cite{li2012code, adel2013recurrent, adel2014features}.

%The role of code-switching in modern societies has been studied by
%many researchers. For example, it was found that code-switching is
%more commonly observed in bilingual or multilingual environment, and
%the acoustic and linguistic change is more significant~\cite{felicity2013mixed}.
%Some other researchers suggest that the gradual fossilisation of code-switching results in
%multiple languages~\cite{auer1999codeswitching, myers2003lies}.

%speech recognition (ASR)~\cite{hinton2012deep, yu2015automatic} recently
%makes it easier to make progress in the field of mixed language.

To promote research on mixlingual speech processing,
the center for
speech and language technologies (CSLT) at Tsinghua
University and SpeechOcean (Beijing Haitian Ruisheng Science Technology Ltd.)
together organize
a Chinese-English mixlingual speech recognition (MixASR-CHEN) challenge on
O-COCOSDA 2016. This event calls for a competition on a speech
recognition task on mixlingual utterances
where one or several English words
are embedded in Chinese utterances. To support this
event, SpeechOcean released a mixlingual speech database
named `OC16-CE80' which consists of $80$ hours of speech signals.
This database is free for the participants of the challenge, and
can be used together with another two free data resources (THCHS30 Chinese speech database and
CMU English dictionary) to build a full-fledged mixlingual speech recognition system.
This paper presents the data profile of the database, the
evaluation rules of the challenge, and a mixlingual ASR baseline system
that the participants can refer to.

Note that there are several other mixlingual databases for different code-switching,
e.g. Cantonese-English, English-Mandarin, Mandarin-Taiwanese and
Mandarin-English~\cite{chan2005development, lyu2006speech, wu2006automatic, lyu2008language, lyu2010analysis}.
The OC16-CE80 database is similar to the SEAME database~\cite{Lyu2015Mandarin} as both are
Mandarin-English code-switching, but the speakers of SEAME are from Singaporean and Malaysian,
whereas the speakers of OC16-CE80 are totally from the China mainland. Due to the less
popularity of spoken English in China compared to Singapore and Malaysia, the code-switching
in OC16-CE80
might be less fluent. Finally, OC16-CE80 consists of $80$ hours of speech signals from nearly $1,500$ speakers,
which is larger than SEAME (157 speakers, 63 hours of signals).

\section{Database profile}
The OC16-CE80 database was originally created by SpeechOcean,
targeting for various mixlingual speech processing tasks (mainly ASR).
The entire database involves about $80$ hours of speech signals,
which were recorded by mobile phones in reading style,
with a sampling rate of $16$ kHz and a sample size of $16$ bits.
The text of the utterances was selected from a large
text corpus, following a greedy search strategy that chose the sentence that
can improve the phone coverage mostly one by one.
The entire database was separated into a training set,
a development set and a test set.
More details of the OC16-CE80 database
are presented in Table~\ref{tab:OC16-CE80}.

\begin{table}[thb!]
\caption{OC16-CE80 Data Profile}
\label{tab:OC16-CE80}
\centerline{
\begin{tabular}{|c|c|c|c|}
\hline
    {\bf Database} & Language & Channel & Utterances/Speaker \\
\hline
    OC16-CE80 & Chinese/English & Mobile & 50 \\
\hline
\hline
    {\bf Datasets} & No. of Speakers & Total Utterances & Recording Hours \\
\hline
    Training set & 1,163 & 58,132 & 63.8 \\
\hline
    Dev. set & 140 & 6,974 & 7.76 \\
\hline
    Test set & 142 &  7,099 & 7.93 \\
\hline
\end{tabular}
}
\begin{tablenotes}
\item[c] Chinese refers to standard Mandarin.
\item[a] Male and Female are mostly balanced.
\item[b] There might be a minor discrepancy with the numbers of total utterances.
\end{tablenotes}
\end{table}

Besides the speech signals,
the OC16-CE80 database also provides the human-labeled transcriptions
for all the recordings of the training set and the development set.
A simple lexicon is also provided to cover most of the commonly used Chinese characters
and the English words appearing in the training and development utterances.
The number of Chinese characters involved
in the lexicon is about $4,653$, with the pronunciations labelled in
tonal Pinyin, e.g., `shang4'.
The number of English words involved in the lexicon is about $6,879$, with the pronunciations excerpted from the
CMU dictionary\footnote{http://svn.code.sf.net/p/cmusphinx/code/trunk/cmudict}.

The OC16-CE80 database is freely available for all the participants
of the MixASR-CHEN challenge and the O-COCOSDA 2016
special session on {\it Mixlingual Speech Processing}. It
is also available for other academic and industrial institutes or
individual users, subject
to a slightly different licence from SpeechOcean.\footnote{\url{http://speechocean.com}}

\section{OC16-MixASR-CHEN challenge}

Based on the OC16-CE80 database, we call a
Chinese-English mixlingual speech recognition (MixASR-CHEN)
challenge.\footnote{\url{http://cslt.riit.tsinghua.edu.cn/mediawiki/index.php/ASR-events-OC16-details}}
The task of the challenge is to train a mixlingual ASR system using the given resources,
and then transcribe the test speech to text using the system within limited time (48 hours).
The performance is measured by word error rate (WER). The evaluation details are described as follows.

\subsection{Development resources}

Three data resources are allowed to utilize in the MixASR-CHEN challenge for system development,
as shown below:

\begin{itemize}
\item OC16-CE80, as described in the previous section. This database contributes the main speech data for acoustic model training. The
transcriptions of the training data are also a good resource for language model training. These transcriptions are not only domain-specific, but also faced with the phenomenon of code-switching.

\item THCHS30~\cite{wang2015thchs}, a pure Chinese speech database provided by CSLT at Tsinghua University,
free and online downloable\footnote{http://www.openslr.org/18}. This
database involves a full set of resources that can be used to develop a full-fledged Chinese ASR system, including
speech signals, transcriptions, a large lexicon and an associated language model. All the resources of THCHS30
can be used in the challenge, but perhaps the most important ones are the lexicon and the language model.

\item CMU English dictionary v0.7b\footnote{\url{http://svn.code.sf.net/p/cmusphinx/code/trunk/cmudict/cmudict-0.7b}}, a
free downloadable resource that can be used to enrich English words in the lexicon.
\end{itemize}

According to the challenge rule, the above three data resources are the only materials that can be used to
develop the recognition system. Any additional resources are not allowed,
including speech, text, lexicon, and supervision from other systems.
However, participants can use their own phone set, in particular for Chinese words.
This is because Chinese pronunciations can be described by either Initial-Finals (IF) or phones,
and neither is more standard than the other one. The Chinese lexicon provided
by THCHS30 is based on IFs, but participants can freely translate them to phones.

\subsection{Test plan}

The MixASR-CHEN challenge chooses WER as the
principle evaluation metric. Note that for Chinese,
we regard each single Chinese character as a word, which
means that the WER on the Chinese part is essentially
the character error rate (CER). Using WER makes the
presentation more clear when the transcriptions involve
both Chinese and English words.

The WER is computed as follows:
\[
WER=\frac{S+D+I}{N},
\]
where $N$ is the total number of words in the
ground-truth transcription, and $S$, $D$, $I$ denote the number of
substitutions, deletions and insertions errors, respectively.
These errors are computed from the forced alignment
between the hypothesized transcriptions and the ground truth.

Note that different participants may treat English words
in different ways. For a fair evaluation, some normalization
is performed before the submitted hypothesis is evaluated.
We first merge scattered letters to
a single word, e.g., `I B M' to `IBM', and then convert
all English words to the upper case.

The challenge reports individual WERs on the Chinese part and English part
of the hypothesis respectively, and the WER on the entire hypothesis.
Although the main metric is the entire WER, the WER on the English part is
also or more important.

\section{Baseline}

We present a baseline MixASR system based on
the deep neural network-hidden Markov model (DNN-HMM) hybrid architecture.
The experiments were conducted with the Kaldi toolkit~\cite{povey2011kaldi}.
The purpose of these experiments is not to present a competitive submission,
but to demonstrate that the OC16-CE80 database is
a reasonable data resource to conduct mixlingual speech recognition
research.

\subsection{Experimental setup}

The speech data used to train the acoustic model was from the OC16-CE80 database
and it contained both training set and development set.
We refrained from using THCHS30 because our focus was to test quality of the new
OC16-CE80 database, and involving additional data may lead to a biased evaluation.
The lexicon used for the baseline system involved two parts:
the Chinese part from the THCHS30 database, and the English part from the CMU English
dictionary v0.7b. The phones of Chinese words and English words were collected, forming a
mixlingual phone set. As it is a baseline system, we did not try to merge phones of different languages.

The acoustic model (AM) of the ASR system was built largely following the Kaldi WSJ s5 nnet3 recipe.
The training started from building an initial Gaussian mixture model-hidden Markov model (GMM-HMM) system,
using Mel frequency cepstral coefficients (MFCCs) as the feature. The initial system involved $4,483$ HMM states and
$3,500$ Gaussian Mixture components (pdfs). This system was employed to produce forced alignment for
the training data, which was used to train the DNN-HMM model.

The DNN model we used was the time-delay neural network (TDNN)~\cite{peddinti2015time}.
It contained $6$ hidden layers, and the activation function was p-norm~\cite{zhang2014improving}.
The input dimension of the p-norm function was $2,000$ and the output dimension was $250$.
%We also spliced the frames for sub-sampling as the recipe suggested.
The natural stochastic gradient descent (NSGD) algorithm was employed
to train the TDNN~\cite{povey2014parallel}.
The input feature involved 40-dimensional Fbanks, with a symmetric $4$-frame window to
splice neighboring frames. The output layer consisted of $3,500$
units, equal to the total number of Gaussian mixtures in the
GMM system that was trained to bootstrap the TDNN model.

As for the language model (LM), we used the conventional 3-grams.
To deal with the mixlingual difficulty, four LM configurations were investigated.
The first one (THLM) was the LM offered by the THCHS30 database.
This LM well matched the lexicon of the baseline system, as the Chinese words in the
lexicon were just from THCHS30. The problem of this LM was that it
involved no English words so it could not handle English at all.
The second LM (OCLM) was trained from the transcriptions of
the training and development data of OC16-CE80. This
LM matched the test domain and involved English words, however
it might suffer from data sparsity.
The third LM (MIX) was a mixture of the above two LMs by a linear interpolation,
where the validation set used to optimize the interpolation factor
was $2,000$ sentences selected from the transcriptions of
the OC16-CE80 database.
In our experiment, we found the interpolation factor for the THLM was
nearly $0.0$, suggesting that THLM does not fit the domain
of the OC16-CE80 database.
The fourth LM (JOIN) was trained with all the transcriptions
of THCHS30 and OC16-CE80. Note that the baseline lexicon was
used when training the OCLM and the JOIN LM, which meant that
all the words in the lexicon were consisted in these LMs,
although the words absent from the training text data
obtained only a small probability, depending on the KN smooth we used
in the experiment.

\subsection{Performance results}

Table~\ref{tab:results} presents the baseline results on the test set, where `THLM', `OCLM', `MIX' and `JOIN'
represent the four LM configurations presented in the previous section.
From the results, we observe that the MIX, which is a mixture of THLM and OCLM and has a high bias on OCLM, works
the best on both Chinese and English words.
The THCHS30 LM is not much suitable for
the OC16-CE80 test: it leads to a high WER on Chinese words, and 100\% WER on English.
Besides, the results of OCLM are not far from those of MIX.
From the above, we know that merging THCHS30 can offer a bit benefit on the LM level.
And the results with the domain-matched LM (OCLM) confirm that a reasonable multilingual
ASR system can be established with the OC16-CE80 database, using the standard modeling
approach. The only problem is that the WER on English words is still high,
which is more clear when compared to the performance on Chinese words.
Nevertheless, the baseline results demonstrate that the OC16-CE80 database is a reasonable
resource for research on mixlingual speech recognition.
Several methods can be simply employed to promote English words, e.g., cross-lingual phone/word sharing
in the AM and LM. The aim of the baseline system is to offer a {\it basic} reference for the
participants and so we do not consider these advanced tricks.

%Then we find that many more errors were generated in English recognizing than in Chinese,
%which indicates improving English recognizing gives more chance to improve the overall performance.

\begin{table}[thb!]
\normalsize
\caption{OC16 MixASR-CHEN Baseline Results}
\label{tab:results}
\centerline{
\begin{tabular}{c|c|c|c}
\hline
                   &\multicolumn{3}{|c}{WRE\%} \\
\hline
    LM             & Chinese & English  & Overall\\
\hline
\hline
    THLM          & 48.38  & 100.00 & 46.33  \\
\hline
    OCLM          & 19.09  & 43.72 & 20.21   \\
\hline
    MIX           &\bf{19.00}  &  \bf{43.67} & \bf{20.09}  \\
\hline
    JOIN          &19.30  &   43.86 & 20.37  \\
\hline
\end{tabular}
}
\end{table}

%We also look into the specific errors made by the system as showed in Table~\ref{tab:results}.

\begin{table}[thb!]
\normalsize
\caption{OC16 MixASR-CHEN Baseline Errors}
\label{tab:results_2}
\centerline{
\begin{tabular}{c|c|c|c}
\hline
                   &\multicolumn{3}{|c}{Error\%} \\
\hline
    Error type            & Chinese  & English & Overall \\
\hline
\hline
    Substitution          & 12.92  & 22.11 & 15.32\\
\hline
    Deletion              & 2.88   & 16.59 & 2.67  \\
\hline
    Insertion             & 3.20   & 4.98 & 2.11  \\
\hline
\end{tabular}
}
\end{table}

In order to gain more insight into the difficulties with mixlingual data,
we look into the three different types of errors with the system using the MIX LM.
The results are showed in Table~\ref{tab:results_2}.
Firstly it can be seen that substitution errors take the largest proportion, as expected.
Secondly we observe that both the substitution and deletion errors
are much more significant for English than for Chinese.
To have a better understanding, we tune the LM weight and word insertion penalty to
balance the insertion and deletion errors, and find that when the performance
on Chinese words improves, the performance on English words decreases, and vice versa.
Further analysis on the resultant transcriptions show that most of the substitution errors overall
on English words occur because some of them are recognized as Chinese words with similar pronunciations.

These observations suggest that the high WER on English words can be attributed to
the competition between the two languages. Firstly, the English phones
in the AM are generally weak, which may result in substitution of
English phones by either the silence phone or the Chinese phones. The former leads to deletions,
and the later leads to substitutions. Secondly, the probabilities of English words in the LM are much
weaker than those of Chinese words, in particular the transition probabilities between English
words and Chinese words. This encourages the decoding to remain in pure Chinese word sequence, and whenever
an English word is encountered, it tends to be skipped over or substituted by a Chinese word, therefore
leading to deletion and substitution errors, respectively. This suggests that a key challenge in
mixlingual ASR is how to deal with the competition among languages, with respect to both phone posteriors
and LM paths.

\section{Conclusion}
We presented the data profile of the OC16-CE80 Chinese-English
mixlingual speech database that
was released to support the MixASR-CHEN challenge on O-COCOSDA
2016. The evaluation rules of the challenge were described, and
a baseline system was presented. We showed that the OC16-CE80
database is a suitable data resource for mixlingual speech recognition
research.

\section*{Acknowledgment}

This work was supported by the National Natural Science Foundation of China under Grant No.61271389 and NO.61371136 and the National Basic Research Program (973 Program) of China under Grant No.2013CB329302.

\newpage

\bibliographystyle{IEEEtran}

\bibliography{mixlingual}

\end{document}